\documentclass[letterpaper]{article} 
\usepackage[draft]{aaai2026}  
\usepackage{times}  
\usepackage{helvet}  
\usepackage{courier}  
\usepackage[hyphens]{url}  
\usepackage{graphicx} 
\usepackage{natbib}  
\usepackage{caption} 
\usepackage{booktabs}
\usepackage{multirow}
\usepackage{amsmath}
\usepackage{algorithm}
\usepackage{algorithmic}

\usepackage{newfloat}
\usepackage{listings}
\DeclareCaptionStyle{ruled}{labelfont=normalfont,labelsep=colon,strut=off} 
\floatstyle{ruled}
\newfloat{listing}{tb}{lst}{}
\floatname{listing}{Listing}
\title{Region-Grounded Vision--Language Learning for Detection-Guided Mammographic Lesion Classification}

\author{
Zhengbo Zhou\textsuperscript{\rm 1},
Jiren Li\textsuperscript{\rm 1},
Dooman Arefan\textsuperscript{\rm 2},
Margarita Zuley\textsuperscript{\rm 2},
Shandong Wu\textsuperscript{\rm 1,2,3}
}

\affiliations{
\textsuperscript{\rm 1}Intelligent Systems Program, University of Pittsburgh, Pittsburgh, PA, USA\\
\textsuperscript{\rm 2}Department of Radiology, University of Pittsburgh, Pittsburgh, PA, USA\\
\textsuperscript{\rm 3}Department of Biomedical Informatics and Bioengineering, University of Pittsburgh, Pittsburgh, PA, USA
}

\begin{document}
\maketitle
\begin{abstract}
Vision–language models trained with contrastive objectives have shown promise in medical image analysis. However, conventional global image–text alignment is ill-suited for mammography, where diagnostically relevant lesions are spatially localized and occupy only a small fraction of the image. Subtle morphological cues critical for malignancy assessment can be diluted when representations are learned at the whole-image level. In this work, we propose a novel region-grounded vision–language learning method for detection-guided mammographic lesion classification. The method mirrors radiologists’ diagnostic paradigm. First, a region–text contrastive pretraining stage aligns lesion-specific features with structured clinical descriptors derived from radiology metadata. To mitigate semantic collapse and background bias in low-vocabulary settings, we introduce a multi-component objective incorporating positive alignment, fine-grained semantic hard negatives, and background suppression. Second, an auxiliary lesion detection head is jointly optimized with contrastive classification to preserve spatial sensitivity and enable localization-aware malignancy classification. Extensive experiments on two independent datasets, CBIS-DDSM and VinDr-Mammo, show superior performance of our method compared to related methods and under in-domain, cross-dataset, and transfer learning settings. 

\end{abstract}


\section{Introduction}

Vision--language models (VLMs) trained with contrastive objectives have shown strong capability in learning transferable visual representations by aligning images with natural language descriptions. In medical imaging, adaptations of CLIP~\cite{radford2021learning}, such as PubMedCLIP~\cite{eslami2023pubmedclip} and BiomedCLIP~\cite{zhang2023biomedclip}, leverage paired image--report data to provide supervision beyond categorical labels. Recently, Mammo-CLIP~\cite{ghosh2024mammo} extended this paradigm to mammography, showing that breast-specific pretraining improves robustness and data efficiency.

However, mammographic diagnosis presents a fundamental challenge to global image--text alignment. In full-field mammograms, diagnostically relevant lesions often occupy less than 2\% of the image area, embedded within large regions of normal parenchyma. Malignancy determination depends on subtle and localized morphological cues, such as margin irregularity, spiculation, and clustered microcalcifications, whose discriminative signal can be easily overwhelmed when representations are learned at the global image level. As a result, conventional CLIP-style alignment risks capturing spurious correlations between overall breast texture and textual descriptions rather than grounding semantics to lesion-specific regions.

In clinical practice, radiologists follow a localization-then-characterization paradigm: they first identify suspicious regions and then evaluate fine-grained lesion morphology to assess malignancy. This diagnostic reasoning suggests that computational models should similarly focus their representational capacity on detected lesion regions rather than on global image embeddings. While region-grounded VLMs such as GLIP~\cite{li2022grounded} and RegionCLIP~\cite{zhong2022regionclip} have demonstrated the value of aligning localized visual features with text in natural images, directly transferring these approaches to mammography is non-trivial. Compared to natural-image datasets, mammographic datasets are significantly smaller, textual metadata is highly structured with limited vocabulary, and visual differences between benign and malignant lesions are subtle and domain-specific. These characteristics introduce three key challenges: (1) Lesion dilution, where global alignment obscures small but diagnostically critical regions; (2) Semantic collapse, where limited and structured metadata increases the risk that contrastive learning captures generic lesion presence rather than fine-grained morphology; and (3) Background bias, where the dominance of normal breast tissue can encourage representations to align textual descriptors with irrelevant background patterns. Addressing these challenges requires a method that explicitly enforces lesion-level semantic grounding while remaining robust in low-data settings.

\begin{figure*}[t]
\centering
\includegraphics[width=\linewidth]{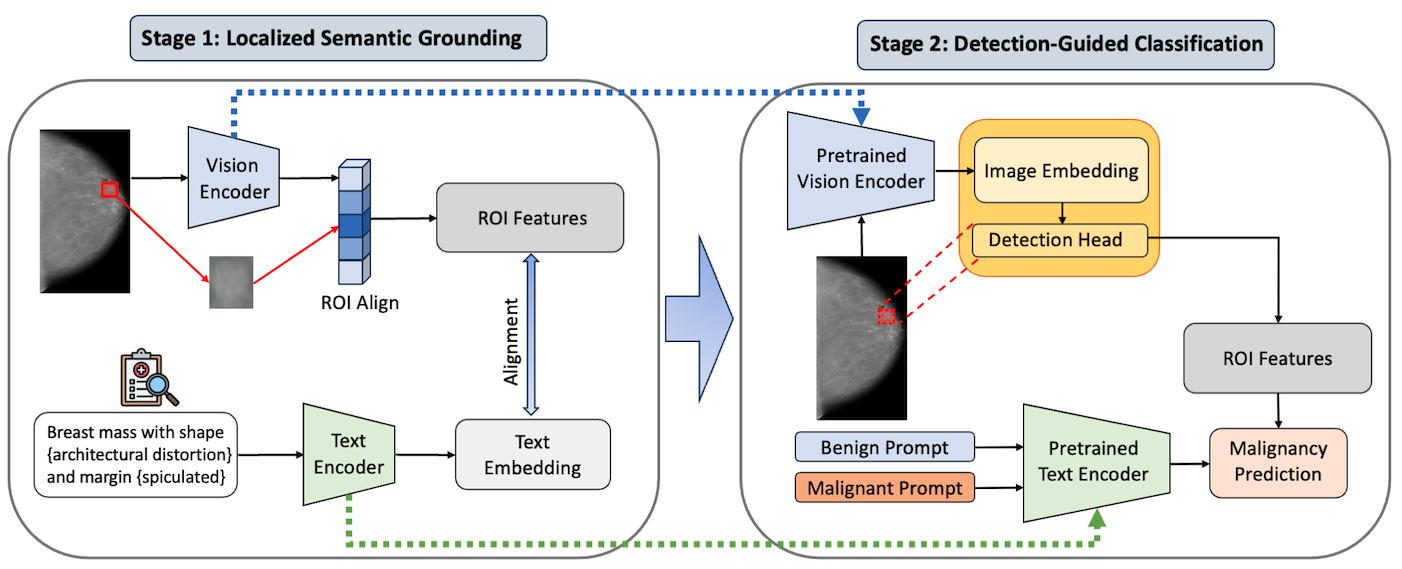}
\caption{Overview of the proposed detection-guided region--text framework.}
\label{fig:framework}
\end{figure*}

In this work, we propose a region--text vision--language learning method for detection-guided mammographic lesion classification. Our method is designed to mirror clinical reasoning and to overcome the limitations of global alignment through targeted objectives. First, we establish localized semantic grounding via region-level contrastive pretraining that aligns lesion-specific ROI features with structured clinical descriptors derived from radiology metadata. To prevent semantic collapse and background bias, we introduce a multi-component objective that combines (i) positive region--text alignment, (ii) fine-grained semantic hard negatives constructed by attribute swapping, and (iii) background region suppression. Second, we incorporate a lightweight FCOS-style auxiliary detection head and jointly optimize detection and contrastive classification, enabling malignancy prediction from lesion-specific representations while preserving spatial sensitivity. A confidence-gated insurance mechanism further improves robustness under uncertain detection.

By explicitly enforcing fine-grained alignment between semantic descriptors and lesion-grounded embeddings, the proposed method preserves subtle morphological features that are critical for malignancy discrimination but often diluted in global embeddings. Extensive experiments on CBIS-DDSM and VinDr-Mammo under in-domain, cross-dataset, and transfer learning settings show consistent improvements over related methods in classification and detection tasks. Our contributions are threefold. 
\begin{itemize} 
\item Region-grounded contrastive learning for mammography: We propose a lesion-level semantic alignment method tailored to the structured and low-vocabulary nature of mammographic metadata. 
\item Multi-component objective to mitigate semantic collapse and background bias: We design semantic hard negatives and background suppression mechanisms specifically for localized clinical semantics. 
\item Detection-guided multimodal learning: We unify auxiliary lesion detection with contrastive classification to preserve spatial sensitivity and improve robustness under domain shift. 
\end{itemize}
This work advances multimodal learning in breast imaging by showing that principled region-level grounding is essential for capturing clinically meaningful morphology in small-lesion imaging settings for breast lesion diagnosis.

\section{Method}

\subsection{Overview}

As shown in Fig. 1, our proposed learning pipeline has a two-stage training curriculum using a shared visual–text backbone. Stage 1 focuses on learning stable localized semantic grounding
between lesion regions and lesion descriptions through a
multi-component contrastive objective. Stage 2 builds on this pretrained representation by introducing a lightweight FCOS-style \cite{tian2019fcos} auxiliary detection head and jointly optimizing detection and classification-related objectives, enabling malignancy prediction from lesion-specific features. At inference, classification is performed by prompt-based similarity scoring using region of interest (ROI) features.

\subsection{Region–Text Contrastive Pretraining}
 We initialize the visual and text encoders with pretrained Mammo-CLIP weights. The visual backbone produces a spatial feature map; we then apply ROIAlign~\cite{he2017mask} to the valid lesion boxes to extract region embeddings. Correspondingly, for each annotated lesion, we construct a clinical description from the associated metadata using a fixed template: \textit{``breast type \{mass/calcification\} with shape \{shape\} and margin \{margin\},''} where the lesion attributes take values from a small, controlled vocabulary.

\subsection{Training Objective}
Stage~1 aims to learn lesion-level semantic grounding from paired region and text embeddings. For each sample, we obtain lesion-region embedding ${\mathbf{v}}$ and one paired text embedding $\mathbf{t}$. We optimize
\begin{equation}
\mathcal{L}_{\text{stage1}}
=
\mathcal{L}_{\text{pos}}
+\lambda_{\text{sem}}\mathcal{L}_{\text{sem-neg}}
+\lambda_{\text{bg}}\mathcal{L}_{\text{bg-neg}},
\label{eq:stage1_loss}
\end{equation}
where $\mathcal{L}_{\text{pos}}$ aligns lesion regions with paired text, $\mathcal{L}_{\text{sem-neg}}$ enforces fine-grained attribute distinction, and $\mathcal{L}_{\text{bg-neg}}$ suppresses spurious alignment to normal tissue.

\noindent\textbf{Positive Contrastive Loss ($\mathcal{L}_{\text{pos}}$).}
Let $\{\mathbf{v}_i\}_{i=1}^{B}$ and $\{\mathbf{t}_i\}_{i=1}^{B}$ denote the region embeddings and paired text embeddings in a mini-batch of size $B$. After $\ell_2$ normalization, we compute the temperature-scaled similarity
$S_{ij}=\mathrm{sim}(\mathbf{v}_i,\mathbf{t}_j)/\tau_{temp}$,
where $\tau_{temp}$ is the temperature parameter. We use a CLIP-style symmetric InfoNCE loss,
$\mathcal{L}_{\text{pos}}=\frac{1}{2}\left(\mathcal{L}_{v\rightarrow t}+\mathcal{L}_{t\rightarrow v}\right)$, with
\begin{equation}
\begin{split}
\mathcal{L}_{v\rightarrow t}
&= -\frac{1}{B}\sum_{i=1}^{B}
\log
\frac{\sum_{j \in \mathcal{P}(i)} \exp(S_{ij})}
{\sum_{j=1}^{B} \exp(S_{ij})}, \\
\mathcal{L}_{t\rightarrow v}
&= -\frac{1}{B}\sum_{i=1}^{B}
\log
\frac{\sum_{j \in \mathcal{Q}(i)} \exp(S_{ji})}
{\sum_{j=1}^{B} \exp(S_{ji})}.
\end{split}
\label{eq:lpos_bidirectional}
\end{equation}
Here, $\mathcal{P}(i)$ and $\mathcal{Q}(i)$ are the sets of valid positive indices for sample $i$ in the region-to-text and text-to-region directions, respectively. In the standard one-positive case, $\mathcal{P}(i)=\mathcal{Q}(i)=\{i\}$. When duplicated structured captions occur due to limited vocabulary within a batch, these sets contain multiple indices, so all matched entries are treated as positives.

\noindent\textbf{Semantic Hard Negative Loss ($\mathcal{L}_{\text{semantic-neg}}$).}
For each lesion sample, we construct mismatched text descriptions by swapping morphological attributes (e.g., shape or margin). We enforce a margin-ranking constraint:
\begin{multline}
\mathcal{L}_{\text{sem-neg}}
=
\frac{1}{B}\sum_{i=1}^{B}\frac{1}{K}\sum_{k=1}^{K}
\max\!\big(0,\; \\
m - \mathrm{sim}(\mathbf{v}_i,\mathbf{t}_i^{+})
+ \mathrm{sim}(\mathbf{v}_i,\mathbf{t}_{i,k}^{-})
\big)
\label{eq:semantic_neg}
\end{multline}
where $\mathbf{t}_i^{+}$ is the matched description embedding, $\mathbf{t}_{i,k}^{-}$ is the $k$-th attribute-swapped description embedding, $K$ is the number of mismatched descriptions, and $m$ is a margin.

\noindent\textbf{Background Negative Loss ($\mathcal{L}_{\text{bg-neg}}$).}
Background regions are sampled as boxes with low IoU to all lesion annotations. Region embeddings $\mathbf{v}^{\text{bg}}_{i,r}$ are extracted via ROIAlign, and we penalize high similarity to the paired lesion text embedding $\mathbf{t}_i^{+}$:
\begin{equation}
\mathcal{L}_{\text{bg-neg}}
=
\frac{1}{B}\sum_{i=1}^{B}\frac{1}{R_i}
\sum_{r=1}^{R_i}
\max\!\left(0,\,
\mathrm{sim}\!\left(\mathbf{v}^{\text{bg}}_{i,r}, \mathbf{t}_i^{+}\right)-\delta
\right),
\label{eq:bg_neg_loss}
\end{equation}
where $R_i$ is the number of sampled background boxes for image $i$ and $\delta$ is a similarity threshold.

\subsection{Detection-Guided Joint Training and Inference}
Stage 2 converts the semantically grounded representation from Stage 1 into a localization-aware classification system by jointly training an auxiliary detector and using features of detected lesions for malignancy prediction. 

\noindent\textbf{Auxiliary Detection Head.}
To encourage the shared encoder to preserve localization-relevant features, we attach a lightweight detection head in an FCOS-style configuration \cite{tian2019fcos}. This head operates on the spatial feature map grid and predicts objectness scores and box regression distances. Benefiting from its anchor-free design, the FCOS-style detector provides lightweight localization supervision with minimal hyperparameter tuning, helping the shared encoder preserve lesion-localized features for ROI-based classification.

\noindent\textbf{Joint Classification and Detection.}
Building on the Stage~1 pretrained encoder, the second stage jointly optimizes the contrastive classification objective and the auxiliary detection head. The total loss becomes:
\begin{equation}
\mathcal{L}_{\text{total}} = \mathcal{L}_{\text{contrastive}} + \mathcal{L}_{\text{det}}.
\label{eq:stage2_loss}
\end{equation}
where $\mathcal{L}_{\text{contrastive}}$ is the symmetric CLIP cross-entropy computed between the ROI-level region embeddings and text embeddings, and $\mathcal{L}_{\text{det}}$ combines focal loss for objectness and smooth L1 loss for bounding box regression from the FCOS-style detection head. 
Classification is performed through prompt-based scoring: for each image, two text prompts (benign/malignant lesion) are encoded, and the class with higher similarity to the region embedding is selected as the prediction. This CLIP-style scoring leverages the region representations learned during both stages without requiring a separate classification head.

\noindent\textbf{Confidence-Gated Insurance on Detection.}
To improve robustness, we use a confidence-gated insurance method to account for the quality of lesion detection. A threshold $\tau$ is used to examine and filter detections with a low confidence from the auxiliary head. A low confidence indicates potentially wrong detections (not related to the lesions). In this case, in order to not be trapped in the wrong detections, we skip ROI-based scoring and instead use the global image embedding $v_{\text{global}}$ from the encoder. This is not ideal but logical because at least the global image contains the lesions, avoiding making classification based on wrongly detected ROIs. The rates of the detection insurance mechanism triggered in our experiments are reported.

\begin{table*}[t]
\centering
\caption{Classification performance of different models under different training settings on the CBIS-DDSM dataset. Calc: Calcification.
Bold: best; underline: second best.}
\label{tab:mass_calc_results}
\begin{tabular}{lcccccc}
\hline
\rule{0pt}{2.6ex} \multirow{2}{*}{\textbf{Model}} &
\multicolumn{3}{c}{\textbf{Mass}} &
\multicolumn{3}{c}{\textbf{Mass + Calc}} \\
\cline{2-4} \cline{5-7}
& \rule{0pt}{2.6ex} Acc & F1 & AUC & Acc & F1 & AUC \\
\hline
\rule{0pt}{2.6ex}DenseNet \cite{iandola2014densenet} & 72.3 & 69.0 & 80.0 & 68.6 & 66.7 & 76.8 \\
ViT \cite{dosovitskiy2020image}& 68.1 & 62.3 & 73.6 & 62.9 & 59.6 & 68.3 \\
FVLM \cite{vo2024frozen} & 76.7 & 74.4 & \underline{84.6} & 70.4 & \underline{67.5} & 78.2 \\
LLaVA-Med \cite{li2023llava} & 72.0 & 71.5 & 82.4 & \underline{71.2} & 66.9 & \underline{78.5} \\
MammoCLIP \cite{ghosh2024mammo} & \underline{77.4} & \underline{74.6} & 83.7 & 71.1 & 66.9 & 77.8 \\
\textbf{Proposed} & \textbf{80.5} & \textbf{75.2} & \textbf{85.5}
                  & \textbf{72.4} & \textbf{69.8} & \textbf{79.2} \\
\hline
\end{tabular}
\end{table*}

\begin{table*}[t]
\centering
\caption{Classification performance of different models under different training settings on the VinDr test set. Bold: best; underline: second best. DDSM denotes CBIS-DDSM.}
\begin{tabular}{lcccccccccccc}
\toprule
\multirow{3}{*}{\textbf{Model}} &
\multicolumn{6}{c}{\shortstack{\strut \textbf{Train: DDSM} \\ \textbf{Test: VinDr}}} &
\multicolumn{6}{c}{\shortstack{\strut \textbf{Train: DDSM+VinDr} \\ \textbf{Test: VinDr}}} \\

\cmidrule(lr){2-7} \cmidrule(lr){8-13}

& \multicolumn{3}{c}{Mass}
& \multicolumn{3}{c}{Mass + Calc}
& \multicolumn{3}{c}{Mass}
& \multicolumn{3}{c}{Mass + Calc} \\

\cmidrule(lr){2-4} \cmidrule(lr){5-7}
\cmidrule(lr){8-10} \cmidrule(lr){11-13}

& Acc & F1 & AUC
& Acc & F1 & AUC
& Acc & F1 & AUC
& Acc& F1 & AUC \\

\midrule
DenseNet \cite{iandola2014densenet}   & 58.6 & 60.6 & 62.7 & 62.3 & \underline{64.9} & 63.3 & 66.3 & 66.9 & 71.5 & 63.5 & 68.5 & 69.6 \\
ViT \cite{dosovitskiy2020image}        & 54.9 & 59.6 & 57.2 & 58.2 & 50.3 & 48.9 & 62.5 & 65.9 & 69.0 & 65.8 & 71.8 & 71.2 \\
FVLM \cite{vo2024frozen}       & \underline{60.7} & 58.7 & \underline{63.3} & 59.7 & 56.5 & 61.8 & 65.8 & 68.9 & 72.1 & 66.2 & 72.9 & \underline{75.0} \\
LLaVA-Med \cite{li2023llava}  & 52.3 & \textbf{68.7} & 53.5 & 59.6 & 27.6 & 53.2 & 62.9 & 67.9 & 72.0 & \underline{68.8} & \textbf{76.7} & 73.7 \\
MammoCLIP \cite{ghosh2024mammo}   & 53.6 & 58.3 & 56.1 & \underline{63.1} & 63.8 & \underline{64.8} & \underline{66.3} & \underline{69.0} & \underline{74.0} & 67.1 & 72.5 & 73.9 \\
\textbf{Proposed}
            & \textbf{61.2} & \underline{67.6} & \textbf{68.5}
            & \textbf{64.0} & \textbf{65.6} & \textbf{66.3}
            & \textbf{73.1} & \textbf{70.4} & \textbf{77.8}
            & \textbf{68.9} & \underline{76.6} & \textbf{78.2} \\
\bottomrule
\end{tabular}
\end{table*}

\begin{table*}[t]
\centering
\caption{Detection performance (mAP) of different models across datasets and training settings. Bold: best; underline: second best. DDSM denotes CBIS-DDSM. Bkg.\ Supp.\ = Background Suppression, Har.\ Neg = Semantic Hard Negatives.}
\label{tab:all_detection}
\setlength{\tabcolsep}{5pt}
\begin{tabular}{lcccccc}
\toprule
\multirow{2}{*}{\textbf{Model}} &
\multicolumn{2}{c}{\shortstack{\strut \textbf{Train: DDSM} \\ \textbf{Test: DDSM}}} &
\multicolumn{2}{c}{\shortstack{\strut \textbf{Train: DDSM} \\ \textbf{Test: VinDr}}} &
\multicolumn{2}{c}{\makebox[0pt]{\shortstack{\strut \textbf{Train: DDSM+VinDr} \\ \textbf{Test: VinDr}}}} \\
\cmidrule(lr){2-3} \cmidrule(lr){4-5} \cmidrule(lr){6-7}
& Mass & Mass+Calc & Mass & Mass+Calc & Mass & Mass+Calc \\
\midrule
YOLOv5 \cite{jocher2022ultralytics} & 49.8 & 21.1 & 14.2 & 11.4 & 55.2 & 34.8 \\
YOLOv12 \cite{tian2025yolov12} & 56.6 & 23.8 & 14.0 & 9.2 & \underline{55.7} & \underline{35.7} \\
MammoCLIP \cite{ghosh2024mammo} & \underline{57.4} & \underline{29.1} & \underline{20.5} & \underline{14.1} & 52.1 & 33.2 \\
\textbf{Proposed} & \textbf{58.1} & \textbf{30.2} & \textbf{20.5} & \textbf{27.2} & \textbf{56.1} & \textbf{40.7} \\
\midrule
W/O Bkg.\ Supp.
& 57.5 & 29.4 & 20.4 & 22.1 & 54.3 & 37.8 \\
W/O Har.\ Neg
& 57.7 & 29.8 & 20.4 & 23.5 & 55.8 & 38.9 \\
W/O Pretrain
& 56.4 & 28.5 & 20.3 & 13.9 & 51.9 & 33.1 \\
\bottomrule
\end{tabular}
\end{table*}

\begin{table*}[t]
\centering
\caption{Ablation studies on classification and comparisons with varying resolutions. Bkg.\ Supp.\ = Background Suppression, Har.\ Neg = Semantic Hard Negatives.}
\begin{tabular}{lcccccccccccc}
\toprule
\multirow{3}{*}{\textbf{Model}} &
\multicolumn{6}{c}{\shortstack{\strut \textbf{Train: DDSM} \\ \textbf{Test: DDSM}}} &
\multicolumn{6}{c}{\shortstack{\strut \textbf{Train: DDSM+VinDr} \\ \textbf{Test: VinDr}}} \\
\cmidrule(lr){2-7} \cmidrule(lr){8-13}
& \multicolumn{3}{c}{Mass} & \multicolumn{3}{c}{Mass + Calc}
& \multicolumn{3}{c}{Mass} & \multicolumn{3}{c}{Mass + Calc} \\
\cmidrule(lr){2-4} \cmidrule(lr){5-7}
\cmidrule(lr){8-10} \cmidrule(lr){11-13}
& Acc & F1 & AUC & Acc & F1 & AUC & Acc & F1 & AUC & Acc & F1 & AUC \\
\midrule

\multicolumn{13}{l}{\textbf{Ablation} ($512\times512$)}\\
\midrule
\textbf{Proposed}
& 80.5 & 75.2 & 85.5
& 72.4 & 69.8 & 79.2
& 73.1 & 70.4 & 77.8
& 68.9 & 76.6 & 78.2 \\
\addlinespace[2pt]
\multicolumn{13}{l}{\emph{Pretrain objectives:}}\\
\quad W/O Bkg.\ Supp.
& 79.1 & 73.8 & 84.2
& 71.8 & 68.2 & 78.1
& 71.8 & 69.9 & 76.5
& 68.4 & 74.1 & 76.8 \\
\quad W/O Har.\ Neg
& 78.3 & 73.0 & 83.9
& 71.4 & 67.3 & 77.8
& 71.2 & 69.5 & 76.1
& 67.9 & 72.5 & 76.2 \\
\quad W/O Pretrain
& 77.9 & 72.6 & 83.4
& 71.2 & 66.6 & 77.2
& 70.9 & 69.8 & 75.8
& 68.1 & 68.9 & 75.2 \\
\addlinespace[2pt]
\multicolumn{13}{l}{\emph{Other components:}}\\
\quad W/O Detection Head
& 76.0 & 71.4 & 82.1
& 68.8 & 66.1 & 74.2
& 69.7 & 67.2 & 74.1
& 65.8 & 71.4 & 74.3 \\
\quad W/O Insurance
& 76.2 & 73.7 & 82.5
& 70.2 & 68.7 & 74.6
& 70.5 & 69.3 & 74.4
& 67.8 & 71.9 & 74.8 \\
\midrule

\multicolumn{13}{l}{\textbf{High-res comparison} ($1520\times912$)}\\
\midrule
\textbf{Proposed}
& 80.7 & 75.5 & 85.7
& 72.5 & 70.3 & 79.8
& 76.3 & 75.2 & 81.2
& 73.8 & 79.3 & 78.9 \\
MammoCLIP \cite{ghosh2024mammo}
& 78.9 & 73.2 & 84.1
& 72.3 & 68.5 & 79.7
& 72.7 & 72.8 & 81.5
& 73.0 & 79.1 & 78.7 \\
\bottomrule
\end{tabular}
\end{table*}

\section{Experiments}

\subsection{Datasets}
Our study received IRB approvals and the main task is breast cancer classification (benign vs.\ malignant) along with lesion detection, on two public mammography datasets: CBIS-DDSM~\cite{sawyer2016curated} and VinDr-Mammo~\cite{Nguyen2022.03.07.22272009}. Lesion labels are based on the originally defined labels of each data set. We analyze mass and calcification separately under two settings: (1) mass-only and (2) mass+calcification. After excluding samples with missing values, CBIS-DDSM contains 1,632 mass cases (861 benign, 771 malignant) and 1,404 calcification cases (737 benign, 667 malignant); VinDr-Mammo contains 1,226 mass cases (530 benign, 696 malignant) and 453 calcification cases (66 benign, 387 malignant). We use the predefined train/test splits originally provided by each dataset.

\subsection{Evaluation Settings.}
We evaluate both \textbf{classification} and \textbf{detection} under the same three settings using full mammograms: (1) \emph{in-domain} (train/test on CBIS-DDSM), (2) \emph{zero-shot cross-dataset} (train on CBIS-DDSM, test on VinDr-Mammo), and (3) \emph{transfer learning} (train on CBIS-DDSM, fine-tune and test on VinDr-Mammo). Unless otherwise specified, images are resized to $512\times512$ and we did a comparison on varying resolutions; for models with pretrained weights, we follow their original pretraining resolution.

\noindent\textbf{Comparisons on classification.}
We compare our proposed model against DenseNet-121~\cite{iandola2014densenet}, ViT-B/16~\cite{dosovitskiy2020image}, Mammo-CLIP~\cite{ghosh2024mammo}, FrozenVLM~\cite{vo2024frozen}, and LLaVA-Med~\cite{li2023llava}. For LLaVA-Med, we apply LoRA~\cite{hu2022lora} for parameter-efficient fine-tuning. For prompt-based scoring, we compute region–text similarities between ROI features and benign/malignant prompts; the predicted label is the higher-scoring prompt. We report Accuracy, F1, and AUC.

\noindent\textbf{Comparisons in detection.}
 We compare against YOLOv5~\cite{jocher2022ultralytics}, YOLOv12~\cite{tian2025yolov12}, and Mammo-CLIP, along with our auxiliary detection head for lesion detection. Detection is evaluated on the test set using score thresholding and non-maximum suppression, reporting AP at IoU$=0.5$ (mAP50) for lesion detection.

\subsection{Experimental Setup}
All models are trained for 50 epochs using AdamW with learning rate $10^{-4}$, weight decay $0.01$, and batch size $8$ on an NVIDIA H100. We set $\tau=0.2$, $\lambda_{\text{sem}}=0.5$, $\lambda_{\text{bg}}=1$, $\delta=0.1$, and $m=0.2$, selected empirically based on validation performance.

\begin{figure*}[t]
\centering
\includegraphics[width=\linewidth]{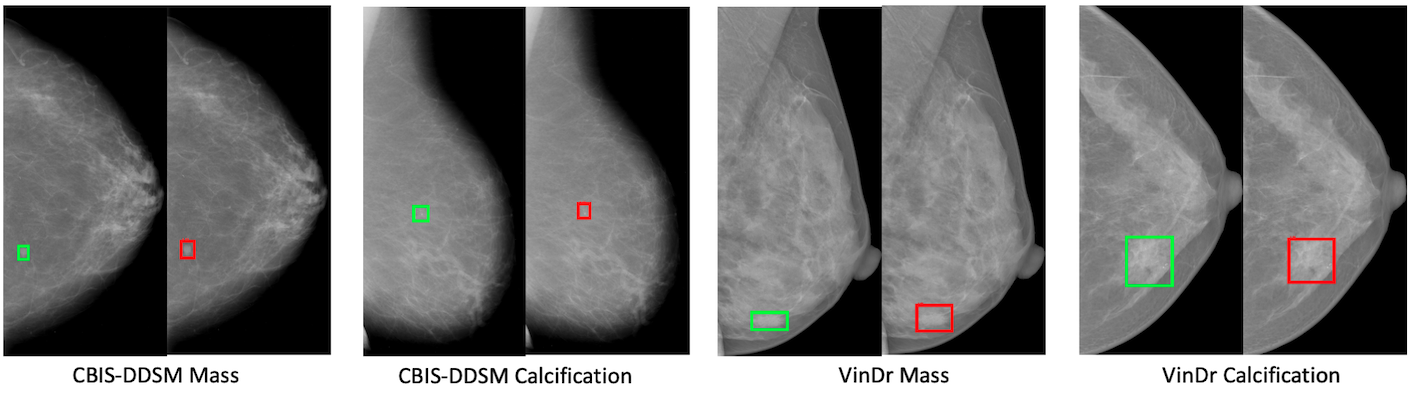}
\caption{Lesion detection visualization. Ground-truth bounding boxes are shown in green (left), and predicted bounding boxes are shown in red (right)}
\label{fig:framework}
\end{figure*}
\section{Results}

\subsection{Classification} In three settings, our method consistently achieves the highest classification performance in CBIS-DDSM (Table~1). Under \emph{zero-shot cross-dataset} evaluation, all methods degrade due to domain shift, yet our approach remains the top performer across lesion settings (Table~2). With \emph{transfer learning}, performance improves and our method again leads, demonstrating strong robustness and transferability.

\subsection{Detection} In all three settings, the proposed method achieves the highest detection performance on both datasets, consistently outperforming YOLOv5, YOLOv12, and MammoCLIP (Table 3).  Compared to the variant trained without Stage~1 pretraining, the proposed model achieves better localization, indicating that Stage~1 strengthens lesion-aware representations and facilitates more accurate lesion detection. Fig.~2 shows several qualitative detection examples, where our model localizes the lesion regions.

\subsection{Ablation Study} As shown in Tables~3 and 4, removing either background suppression or semantic hard negatives individually degrades both detection and classification performance, confirming their complementary contributions — the former enhances localization precision, while the latter strengthens semantic discriminability through contrastive alignment. 

Omitting Stage~1 pretraining entirely yields a compounded decline across all settings. Ablating the detection head incurs the largest classification drop, as the alignment objectives lose access to well-localized region proposals and are forced to operate on global image representations. Finally, removing confidence-gated insurance particularly degrades multi-class performance, as unreliable detections propagate to downstream classification.

Across all experiments, the detection insurance mechanism is triggered for 3\%--8\% of all testing cases, indicating lesion localization has a low confidence. Higher-resolution evaluation yields modest gains, and our method still outperforms MammoCLIP under the same resolution setting.

\section{Discussion and Conclusion}
We proposed a novel region--text contrastive learning framework with localized semantic 
grounding for detection-guided mammographic lesion classification. The fine-grained 
alignment between semantic textual descriptors and region-grounded lesion embeddings 
enhances lesion diagnosis by preserving subtle morphological features, such as margin 
irregularity, spiculation, and microcalcification patterns, that are otherwise diluted 
by the dominant normal parenchymal tissue in full-size mammograms where lesions occupy 
only a small fraction of the image. By anchoring contrastive objectives to 
detection-guided region proposals, our framework ensures that learned representations 
capture clinically discriminative information rather than global tissue-level statistics. 
Complementary training mechanisms, including semantic hard negatives and background 
suppression, further prevent representation collapse and mitigate background bias.

Extensive experiments on two independent mammogram datasets demonstrate superior 
performance compared to state-of-the-art methods across multiple evaluation metrics. 
Comprehensive ablation studies confirm the contribution of each proposed component, 
validating that principled region-level grounding is essential for capturing clinically 
meaningful morphology in small-lesion imaging settings. 

In the future, we will explore extending 
this framework to other imaging modalities where lesion-to-image size ratios present 
similar challenges, and investigating the integration of longitudinal temporal 
information for improved diagnostic accuracy.

\section{ACKNOWLEDGMENTS}

This work was supported in part by a NIH Other Transaction research contract \#1OT2OD037972-01 and \#3OT2OD037972-01S1, the grant 1R01EB032896 as part of the NSF/NIH Smart Health and Biomedical Research in the Era of Artificial Intelligence and Advanced Data Science Program, an Amazon Machine Learning Research Award, a PA Breast Cancer Coalition grant, a Jewish Healthcare Foundation grant, and the University of Pittsburgh Momentum Funds (Scaling Grant) for the Pittsburgh Center for AI Innovation in Medical Imaging. This work used Bridges-2 at the Pittsburgh Supercomputing Center through allocation [MED200006] from the Advanced Cyberinfrastructure Coordination Ecosystem: Services \& Support (ACCESS) program, which is supported by NSF grants \#2138259, \#2138286, \#2138307, \#2137603, and \#2138296. This research was also supported in computing resources by the University of Pittsburgh Center for Research Computing and Data (RRID: SCR\_022735) through the resources provided by the H2P cluster, which is supported by NSF award OAC-2117681. The views and conclusions contained in this document are those of the authors and should not be interpreted as representing official policies, either expressed or implied, of the NIH or NSF.

\bibliography{aaai2026}

\end{document}